# Assessment and Linear Programming Under Fuzzy Conditions

**Michael Voskoglou***
**Department of Mathematical Sciences, University of Peloponnese, Graduate TEI of Western Greece.**

**A B S T R A C T**

**The present work focuses on two directions. First, a new fuzzy method using triangular/trapezoidal fuzzy numbers as tools is developed for evaluating a group's mean performance, when qualitative grades instead of numerical scores are used for assessing its members' individual performance. Second, a new technique is applied for solving Linear Programming problems with fuzzy coefficients. Examples are presented on student and basket-ball player assessment and on real life problems involving Linear Programming under fuzzy conditions to illustrate the applicability of our results in practice. A discussion follows on the perspectives of future research on the subject and the article closes with the general conclusions.**

## 1. Introduction

Despite to the initial reserve of the classical mathematicians, fuzzy mathematics and Fuzzy Logic (FL) have found nowadays many and important applications to almost all sectors of human activity (e.g. [1], Chapter 6, [2] Chapters 4-8, [3], etc.). Due to its nature of characterizing the ambiguous real life situations with multiple values, FL offers among others rich resources for assessment purposes, which are more realistic than those of the classical logic [4-6].

Fuzzy Numbers (FNs), which are a special form of Fuzzy Sets (FS) on the set of real numbers, play an important role in fuzzy mathematics analogous to the role played by the ordinary numbers in the traditional mathematics. The simplest forms of FNs are the Triangular FNs (TFNs) and the Trapezoidal FNs (TpFNs).

In the present work we study applications of TFNs and TpFNs to assessment processes and to Linear Programming (LP) under fuzzy conditions. The rest of the paper is formulated as follows: Section 2 contains all the information about FS, FNs and LP which is necessary for the understanding of its contents. Section 3 is divided in two parts. In the first part an assessment method is developed using TFNs/TpFNs as tools, which enables the calculation of the mean performance of a group of uniform objects (individuals, computer systems, etc.) with respect to a common activity performed under fuzzy conditions. In the second part a method is developed for solving LP problems with fuzzy data (Fuzzy LP). In Section 4 examples are presented illustrating the applicability of both methods to real world situations. The assessment outcomes are validated with the parallel use of the GPA index, while the solution of the FLP problems is reduced to the solution of ordinary LP problems by ranking the corresponding fuzzy coefficients. The article closes with a brief discussion for the perspectives of future research on those topics and the final conclusions that are presented in Section 5.

* Corresponding author
E-mail address: mvoskoglou@gmail.com

## 2. Background

### 2.1. Fuzzy Sets and Logic

For general facts on FS and FL we refer to [2] and for more details to [1]. The FL approach for a problem's solution involves the following steps:

- Fuzzification of the problem's data by representing them with properly defined FSs.
- Evaluation of the fuzzy data by applying principles and methods of FL in order to express the problem's solution in the form of a unique FS.
- Defuzzification of the problem's solution in order to "translate" it in the natural language for use with the original real-life problem.

One of the most popular defuzzification methods is the Center of Gravity (COG) technique. When using it, the fuzzy outcomes of the problem's solution are represented by the coordinates of the COG of the membership function graph of the FS involved in the solution [7].

### 2.2. Fuzzy Numbers

It is recalled that a FN is defined as follows:

**Definition 1.** A FN is a FS A on the set $R$ of real numbers with membership function

$m_A: R \to [0, 1]$, such that:

- A is normal, i.e. there exists $x$ in $R$ such that $m_A(x) = 1$,
- A is convex, i.e. all its $a$-cuts $A^a = \{x \in U: m_A(x) \geq a\}$, $a$ in [0, 1], are closed real intervals, and
- Its membership function $y = m_A(x)$ is a piecewise continuous function.

**Remark 1.** (Arithmetic operations on FNs)**:** One can define the four basic arithmetic operations on FNS in the following two, equivalent to each other, ways:

- With the help of their a-cuts and the Representation-Decomposition Theorem of Ralesscou - Negoita ([8], Theorem 2.1, p.16) for FS. In this way the fuzzy arithmetic is turned to the well known arithmetic of the closed real intervals.

– By applying the Zadeh's extension principle ([1], Section 1.4, p.20), which provides the means for any function f mapping a crisp set X to a crisp set Y to be generalized so that to map fuzzy subsets of X to fuzzy subsets of Y.

In practice the above two general methods of the fuzzy arithmetic, requiring laborious calculations, are rarely used in applications, where the utilization of simpler forms of FNs is preferred. For general facts on FNs we refer to [9].

## 2.3. Triangular Fuzzy Numbers (TFNs)

A TFN (*a, b, c*), with *a, b, c* in *R* represents mathematically the fuzzy statement "the value of *b* lies in the interval [*a, c*]". The membership function of (*a, b, c*) is zero outside the interval [*a, c*], while its graph in [*a, c*] consists of two straight line segments forming a triangle with the OX axis (***Fig.1***).

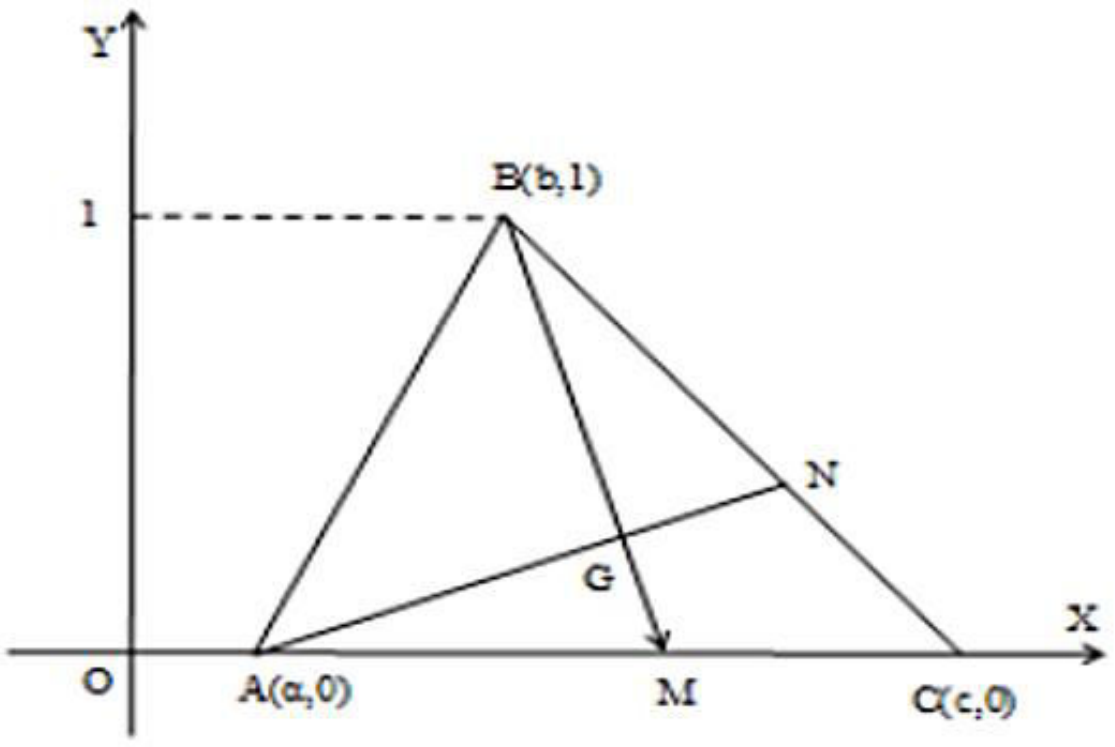

***Fig. 1.*** Graph and COG of the TFN (a, b, c).

Therefore the analytic definition of a TFN is given as follows:

**Definition 2.** Let *a, b* and *c* be real numbers with $a < b < c$. Then the *TFN* (*a*, *b*, *c*) is a FN with membership function:

$$y = m(x) = \begin{cases} \frac{x-a}{b-a}, & x \in [a,b] \\ \frac{c-x}{c-b}, & x \in [b,c] \\ 0, & x < a \text{ or } x > c \end{cases}$$

**Proposition 1**. (Defuzzification of a TFN). The coordinates (*X*, *Y*) of the COG of the graph of the TFN (*a*, *b*, *c*) are calculated by the formulas $X = \frac{a+b+c}{3}$, $Y = \frac{1}{3}$.

**Proof.** The graph of the TFN (*a*, *b*, *c*) is the triangle ABC of ***Fig.1***, with A (*a*, 0), B(*b*, 1) and C (*c*, 0). Then, the COG, say G, of ABC is the intersection point of its medians AN and BM. The proof of the proposition is easily obtained by calculating the equations of AN and BM and by solving the linear system of those two equations.

**Remark 2.** (Arithmetic Operations on TFNs) It can be shown [9] that the two general methods of defining arithmetic operations on FNs mentioned in ***Remark 2*** lead to the following simple rules for the addition and subtraction of TFNs:

Let A = ($a, b, c$) and B = ($a_1, b_1, c_1$) be two TFNs. Then:

- The sum A + B = ($a+a_1, b+b_1, c+c_1$).
- The difference A - B = A + (-B) = ($a-c_1, b-b_1, c-a_1$), where –B = ($-c_1, -b_1, -a_1$) is defined to be the opposite of.

In other words, the opposite of a TFN, as well as the sum and the difference of two TFNs are always TFNs. On the contrary, the product and the quotient of two TFNs, although they are FNs, they are not always TFNs, unless if $a, b, c, a_1, b_1, c_1$ are in $R^+$ [9].

The following two scalar operations can be also be defined:

- k + A= (k+$a$, k+$b$, k+$c$), $k \in R$.
- kA = (k$a$, k$b$, k$c$), if k>0 and kA = (k$c$, k$b$, k$a$), if k<0.

## 2.4. Trapezoidal Fuzzy Numbers (TpFNs)

A TpFN ($a, b, c, d$) with $a, b, c, d$ in $R$ represents the fuzzy statement approximately in the interval $[b, c]$. Its membership function $y=m(x)$ is zero outside the interval [$a, d$], while its graph in this interval [$a, d$] is the union of three straight line segments forming a trapezoid with the X-axis (see ***Fig. 2***),

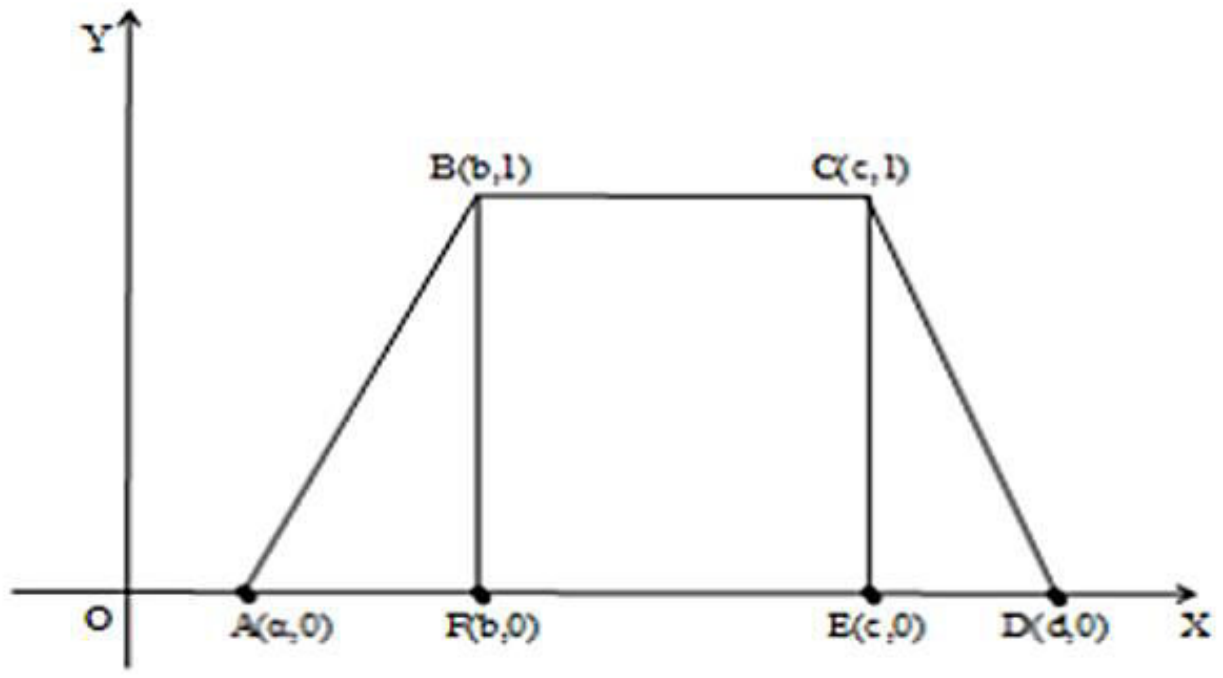

***Fig. 2.*** Graph of the TpFN (a, b, c, d).

Therefore, the analytic definition of a TpFN is given as follows:

**Definition 3.** Let $a < b < c < d$ be given real numbers. Then the TpFN ($a, b, c, d$) is the FN with membership function:

$$
y = m(x) = \begin{cases} \frac{x-a}{b-a} & , \quad x \in [a,b] \\ x = 1, & , \quad x \in [b,c] \\ \frac{d-x}{d-c}, & \quad x \in [c,d] \\ 0, & \quad x < a \text{ and } x > d \end{cases}
$$

**Remark 3.**

- It is easy to observe that the TpFNs are generalizations of TFNs. In fact, the TFN (*a, b, d*) can be considered as a special case of the TpFN (*a, b, c, d)* with *b=c*.
- The TFNs and the TpFNs are special cases of the *LR – FNs* of Dubois and Prade [10]. Generalizing the definitions of TFNs and TpFNs one can define n-agonal FNs of the form ($a_1$, $a_2, \ldots , a_n$) for any integer *n*, $n \geq 3$; e.g. see Section 2 of [11] for the definition of the hexagonal FNs.
- It can be shown [9] that the addition and subtraction of two TpFNs are performed in the same way that it was mentioned in ***Remark 2*** for TFNs. Also, the two scalar operations that have been defined in ***Remark 2*** for TFNs hold also for TpFNs.

The following two propositions provide two alternative ways for defuzzifying a given TpFN:

**Proposition 2.** (GOG of a TpFN)**:** The coordinates (*X*, *Y*) of the COG of the graph of the TpFN (*a, b, c, d*) are calculated by the formulas $X = \frac{c^2+d^2-a^2-b^2+dc-ba}{3(c+d-a-b)}$, $Y = \frac{2c+d-a-2b}{3(c+d-a-b)}$.

**Proof.** We divide the trapezoid forming the graph of the TpFN (*a, b, c, d*) in three parts, two triangles and one rectangle (***Fig. 2***). The coordinates of the three vertices of the triangle ABE are (*a,* 0), (*b,* 1) and (*b,* 0) respectively, therefore by ***Proposition 1*** the COG of this triangle is the point $C_1\ (\frac{a+2b}{3}, \frac{1}{3})$.

Similarly one finds that the COG of the triangle FCD is the point $C_2\ (\frac{d+2c}{3}, \frac{1}{3})$. Also, it is easy to check that the COG of the rectangle BCFE, being the intersection point of its diagonals, is the point $C_3\ (\frac{b+c}{2}, \frac{1}{2})$. Further, the areas of the two triangles are equal to $S_1 = \frac{b-a}{2}$ and $S_2 = \frac{d-c}{2}$ respectively, while the area of the rectangle is equal to $S_3 = c- b$.

Therefore, the coordinates of the COG of the trapezoid, being the resultant of the COGs $C_i$ ($x_i$, yi), for i=1, 2, 3, are calculated by the formulas $X = \frac{1}{S}\sum_{i=1}^{3} S_i x_i$, $Y = \frac{1}{S}\sum_{i=1}^{3} S_i y_i$ (1), where $S = S_1 + S_2 + S_3 = \frac{c+d-b-a}{2}$ is the area of the trapezoid [12].

The proof is completed by replacing the above found values of $S$, $S_i$, $x_i$ and $y_i$, i = 1, 2, 3, in formulas (1) and by performing the corresponding operations.

**Proposition 3.** (GOG of the GOGs of a TpFN)**:** Consider the graph of the TpFN ($\alpha$, $b$, $c$, $d$) (***Fig. 3***). Let $G_1$ and $G_2$ be the COGs of the rectangular triangles AEB and CFD and let $G_3$ be the COG of the rectangle BEFC respectively. Then $G_1G_2G_3$ is always a triangle, whose COG has coordinates

$$X = \frac{2(a+d)+7(b+c)}{18},\ Y = \frac{7}{18}.$$

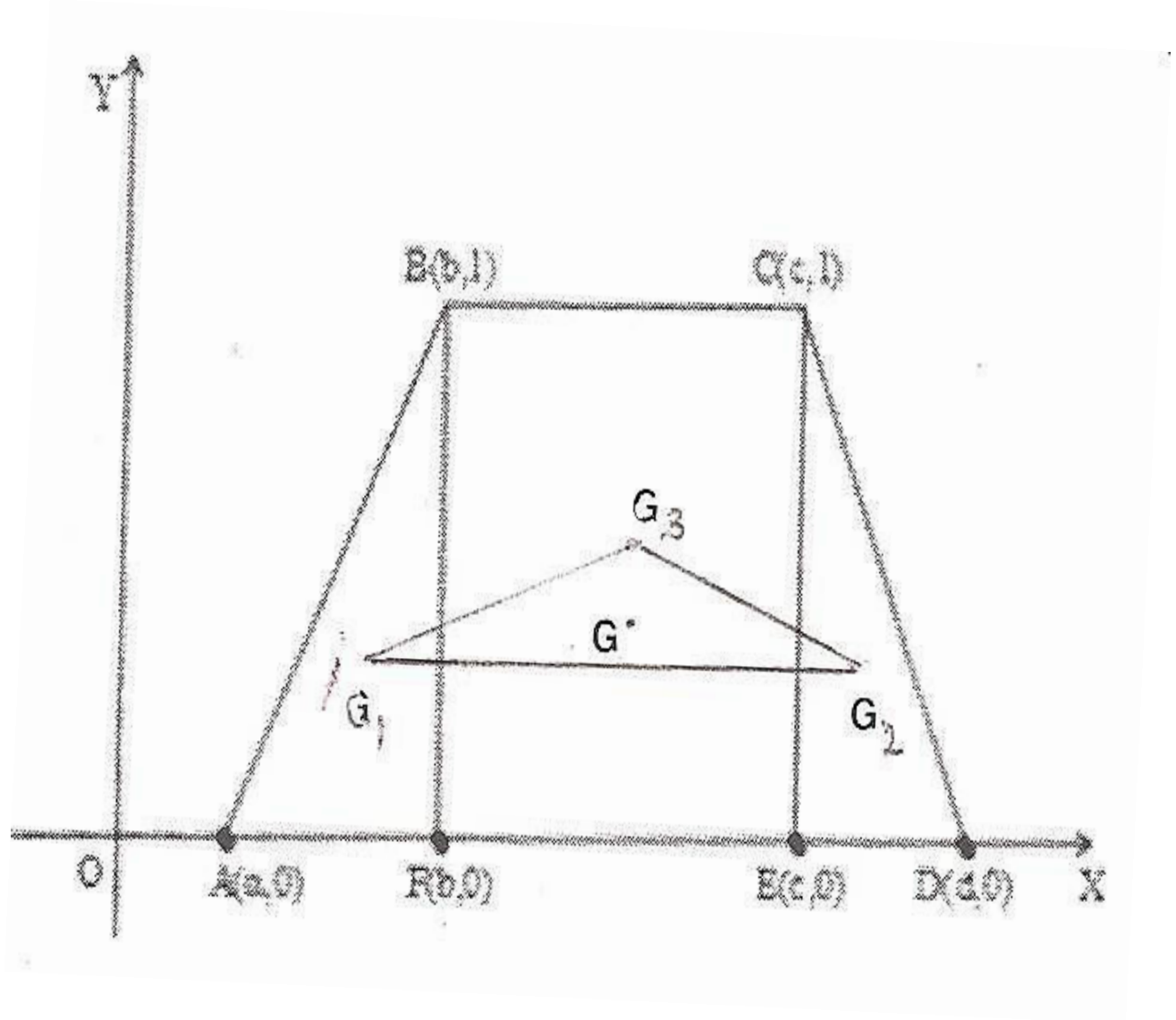

***Fig. 3.*** The GOG of the GOGs of the TpFN (a, b, c, d).

**Proof.** By ***Proposition 1*** one finds that $G_1\left(\frac{a+2b}{3},\frac{1}{3}\right)$ and $G_2\left(\frac{d+2c}{3},\frac{1}{3}\right)$. Further, it is easy to check that the GOG $G_3$ of the rectangle BCFD, being the intersection of its diagonals, has coordinates $\left(\frac{b+c}{2},\frac{1}{2}\right)$.

The y – coordinates of all points of the straight line defined by the line segment $G_1G_2$ are equal to $\frac{1}{3}$, therefore the point $G_3$, having y – coordinate equal to $\frac{1}{2}$, does not belong to this line. Hence, by ***Proposition 1***, the COG $G'$ of the triangle $G_1G_2G_3$ has coordinates $X = \left(\frac{a+2b}{3}+\frac{d+2c}{3}+\frac{b+c}{2}\right): 3 = \frac{2(a+d)+7(b+c)}{18}$ and $Y = \left(\frac{1}{3}+\frac{1}{3}+\frac{1}{2}\right): 3 = \frac{7}{18}$.

**Remark 4.** Since the COGs $G_1$, $G_2$ and $G_3$ are the balancing points of the triangles AEB and CFD and of the rectangle BEFC respectively, the COG $G'$ of the triangle $G_1G_2G_3$, being the balancing point of the triangle formed by those COGs, can be considered instead of the COG G of the trapezoid ABCD for defuzzifying the TpFN ($a$, $b$, $c$, $d$). The advantage of the choice of $G'$ instead of G is that the formulas calculating the coordinates of $G'$ (***Proposition 3***) are simpler than those calculating the COG G of the trapezoid ABCD (***Proposition 2***).

An important problem of the fuzzy arithmetic is the ordering of FNs, i.e. the process of determining whether a given FN is larger or smaller than another one. This problem can be solved through the introduction of a ranking function, say R, which maps each FN on the real line, where a natural order exists. Several ranking methods have been proposed until today, like the lexicographic screening [13], the use of an area between the centroid and original points [14], the subinterval average method [11], etc.

Here, under the light of ***Propositions (1) & (3)*** and of ***Remark 4***, we define the ranking functions for TFNs and TpFNs as follows:

**Definition 4.** Let *A* be a FN. Then:

- If *A* is a TFN of the form *A* {*α, b, c*}, we define $R(A) = \frac{a+b+c}{3}$.
- If *A* is a TpFN of the form *A* {α, *b, c, d)*, we define $R(A) = \frac{2(a+d)+7(b+c)}{18}$.

## 2.5. Linear Programming

It is well known that Linear Programming (LP) is a technique for the optimization (maximization or minimization) of a linear objective function subject to linear equality and inequality constraints. The feasible region of a LP problem is a convex polytope, which is a generalization of the three-dimensional polyhedron in the *n*-dimensional real space $R^n$, where *n* is an integer, $n \geq 2$.

A LP algorithm determines a point of the LP polytope, where the objective function takes its optimal value, if such a point exists. Dantzig invented the SIMPLEX algorithm [15] that has efficiently tackled the LP problem in most cases. Further, Dantzig, adopting a conjecture of John von Neuman, who worked on an equivalent problem in Game Theory, provided a formal proof of the theory of Duality [16]**.** According to the above theory every LP problem has a dual problem the optimal solution of which, if there exists, provides an optimal solution of the original problem. For general facts about the SIMPLEX algorithm we refer to Chapters 3 and 4 of [17].

LP, apart from mathematics, is widely used nowadays in business and economics, in several engineering problems, etc. Many practical problems of operations research can be expressed as LP problems. However, in large and complex systems, like the socio-economic, the biological ones, etc. ., it is often very difficult to solve satisfactorily the LP problems with the standard theory, since the necessary data cannot be easily determined precisely and therefore estimates of them are used in practice. The reason for this is that such kind of systems usually involve many different and constantly changing factors the relationships among which are indeterminate, making their operation mechanisms to be not clear. In order to obtain good results in such cases one may apply techniques FLP, e.g. see [18, 19], etc.

## 3. Main Results

### 3.1. Assessment under Fuzzy Conditions

Assume that one wants to evaluate the mean performance of a group of uniform objects (individuals, computer systems, etc.) participating in a common activity. When the individual performance of the group's members is assessed by using numerical grades (scores), the traditional method for evaluating the group's mean performance is the calculation of the mean value of those scores. However, cases appear frequently in practice, where the individual performance is assessed by using linguistic instead of numerical grades. For example, this frequently happens for student assessment, usually for reasons of more elasticity that reduces the student pressure created by the existence of the strict numerical scores.

A standard method for such kind of assessment is the use of the linguistic expressions (labels) A = excellent, B = very good, C = good, D = fair and F = unsatisfactory (failed). In certain cases some insert the label E = less than satisfactory between D and F, while others use labels like $B^+$, $B^-$, etc., for a more strict assessment. It becomes evident that such kind of assessment involves a degree of fuzziness caused by the existence of the linguistic labels, which are less accurate than the numerical scores. Obviously, in the linguistic assessment the calculation of the mean value of the group's members' grades is not possible.

An alternative method for assessing a group's overall performance in such cases is the calculation of the Grade Point Average (GPA) index ([2], Chapter 6, p.125). The GPA index is a weighted mean calculated by the formula

$$GPA = \frac{0n_F + 1n_D + 2n_C + 3n_B + 4n_A}{n} \quad . \tag{1}$$

In the above formula $n$ denotes the total number of the group's members and $n_A$, $n_B$, $n_C$, $n_D$ and $n_F$ denote the numbers of the group's members that demonstrated excellent, very good, good, fair and unsatisfactory performance respectively. In case of the ideal performance ($n_A = n$) formula (1) gives that GPA = 4, whereas in case of the worst performance ($n_F = n$) it gives that $GPA = 4$. Therefore, we have in general that $0 \leq GPA \leq 4$, which means that values of GPA greater than 2 could be considered as indicating a more than satisfactory performance. However, since in ***Eq. (1)*** greater coefficients (weights) are assigned to the higher scores, it becomes evident that the GPA index reflects actually not the mean, but the group's quality performance.

Here a method will be developed for an approximate evaluation in such cases of the group's mean performance that uses TFNs or TpFNs as tools. For this, we need the following definition:

**Definition 5.** Let $A_i = (a_{1i}, a_{2i}, a_{3i}, a_{4i})$, i = 1, 2,…, $n$ be TFNs/TpFNs (see ***Remark 3***), where $n$ is a nonnegative integer, $n \geq 2$. Then the mean value of the $A_i$'s is defined to be the TFN /TpFN,

$$A = \frac{1}{n}(A_1 + A_2 + \ldots + A_n). \qquad (2)$$

In case of utilizing TFNs as tools the steps of the new assessment method are the following:

- Assign a scale of numerical scores from 1 to 100 to the linguistic grades A, B, C, D and F as follows: A (85–100), B (75–84), C (60–74), D (50–59) and F (0–49)[1].
- For simplifying the notation use the same letters to represent the above grades by the TFNs.
- A = (85, 92.5, 100), B = (75, 79.5, 84), C (60, 67, 74), D (50, 54.5, 59) and F (0, 24.5, 49), respectively, where the middle entry of each of them is equal to the mean value of its other two entries.
- Evaluate the individual performance of all the group's members using the above qualitative grades. This enables one to assign a TFN A, B, C, D or F to each member. Then the mean value M of all those TFNs is equal to the TFN

  $$M(a, b, c) = \frac{1}{n}(n_A A + n_B B + n_C C + n_D D + n_F F).$$

- Use the TFN M (a, b, c) for evaluating the group's mean performance. It is straightforward to check that the three entries of the TFN M are equal to

  $$a = \frac{85n_A + 75n_B + 60n_C + 50n_D + 0n_F}{n}, \quad b = \frac{92.5n_A + 79.5n_B + 67n_C + 54.5n_D + 24.5n_F}{n} \text{ and}$$

  $$c - \frac{100n_A + 84n_B + 74n_C + 59n_D + 49n_F}{n}.$$ Then, by ***Proposition 1*** one gets that

  $$X(M) = \frac{a+b+c}{3} = \frac{92.5n_A + 79.5n_B + 67n_C + 54.5n_D + 24.5n_F}{n} = \frac{a+c}{2} = b \quad (3).$$

- Observe that, in the extreme (hypothetical) case where the lowest possible score has been assigned to each member of the group (i.e. the score 85 to nA members, the score 75 to nB members, etc.) the mean value of all those scores is equal to a. On the contrary, if the greatest score has been assigned to each member, then the mean value of all scores is equal to c. Therefore the value of X(M), being equal to the mean value of a and c, provides a reliable approximation of the group's mean performance.

In cases where multiple referees of the group's performance exist one could utilize TpFNs instead of TFNs for evaluating the group's mean performance. In that case a different TpFN is assigned to each member of the group representing its individual performance, while the other steps of the method remain unchanged (see ***Example 2***).

---

[1]The scores assigned to the linguistic grades are not standard and may differ from case to case. For instance, in a more rigorous assessment one could take A(90-100), B (80-89), C(70-79), D (60-69), F(<60), etc.

### 3.2. Fuzzy Linear Programming

The general form of a FLP problem is the following: Maximize (or minimize) the linear expression

$F = A_1x_1 + A_2x_2 + .... + A_nx_n$ subject to constraints of the form $x_j \geq 0$,

$A_{i1}x_1 + A_{i2}x_2 + ..... + A_{in}x_n \leq (\geq) B_i$, where $i = 1, 2, ..., m$, $j = 1, 2,,,, n$ and $A_j$, $A_{ij}$, $B_i$ are FNs. Here a new method will be proposed for solving FLP problems. We start with the following definition:

**Definition 5.** The *Degree of Fuzziness (DoF)* of a n-agonal FN $A = (a_1, ... , a_n)$ is defined to be the real number $D = a_n - a_1$. We write then $DoF(A) = D$.

The following two propositions are needed for developing the new method for solving FLP problems:

**Proposition 4.** Let $A$ be a TFN with $DoF(A) = D$ and $R(A) = R$. Then $A$ can be written in the form

A = (α, 3R-2α-D, α + D), where $\alpha$ is a real number such that $R - \frac{2D}{3} < \alpha < R - \frac{D}{3}$.

**Proof.** Let $A (\alpha, b, c)$ be the given TFN, with $\alpha, b, c$ real numbers such that $\alpha < b < c$. Then, since

$DoF(A) = c - \alpha = D$, is $c = \alpha + D$. Therefore, $R(A) = \frac{a+b+c}{3} = \frac{2a+b+D}{3} = R$, which gives that

$b = 3R-2\alpha-D$. Consequently we have that $\alpha < 3R-2\alpha-D < \alpha + D$. The left side of the last inequality implies that $3\alpha < 3R-D$, or $\alpha < R-\frac{D}{3}$. Also its right side implies that $-3\alpha < 2D-3R$, or $\alpha > R-\frac{2D}{3}$, which completes the proof.

**Proposition 5.** Let $A$ be a TpFN with $DoF(A) = D$ and $R(A) = R$. Then $A$ can be written in the form

$A = (\alpha, b, c, \alpha + D)$, where $\alpha$, $b$ and $c$ are real numbers such that $\alpha < b \leq c < a + D$ and

$b + c = \frac{18R-4a-2D}{7}$.

**Proof.** Let $A (\alpha, b, c, d)$ be the given TFN, with $\alpha, b, c, d$ real numbers such that $a < b \leq c < d$. Since $D(A) = d - \alpha = D$, it is $d = \alpha + D$. Also, by ***Definition 4*** we have that $R = \frac{2(2a+D)+7(b+c)}{18}$ wherefrom one gets the expression of $b + c$ in the required form.

The proposed in this work method for solving a Fuzzy LP problem involves the following steps:

- Ranking of the FNs $A_j$, $A_{ij}$ and $B_i$.
- Solution of the obtained in the previous step ordinary LP problem with the standard theory.
- Conversion of the values of the decision variables in the optimal solution to FNs with the desired DoF.

The last step is not compulsory, but it is useful in problems of vague structure, where a fuzzy expression of their solution is often preferable than the crisp one (see ***Examples 2, 3***).

# 4. Applications

## 4.1. Examples of Assessment Problems

**Example 1.** ***Table 1*** depicts the performance of students of two Departments, say $D_1$ and $D_2$, of the School of Management and Economics of the Graduate Technological Educational Institute (T. E. I.) of Western Greece in their common progress exam for the course "Mathematics for Economists I" in terms of the linguistic grades A, B, C, D and F:

***Table 1.*** Student performance in terms of the linguistic grades.

| Grade | $D_1$ | $D_2$ |
|---|---|---|
| **A** | 60 | 60 |
| **B** | 40 | 90 |
| **C** | 20 | 45 |
| **D** | 30 | 45 |
| **F** | 20 | 15 |
| **Total** | 170 | 255 |

It is asked to evaluate the two Departments overall quality and mean performance.

- Quality performance (GPA index): Replacing the data of ***Table 1*** to formula (1) and making the corresponding calculations one finds the same value $GPA = \frac{43}{17} \approx 2.53$ for the two Departments that indicates a more than satisfactory quality performance.
- Mean performance (using TFNs): According to the assessment method developed in the previous section it becomes clear that Table 1 gives rise to 170 TFNs representing the individual performance of the students of $D_1$ and 255 TFNs representing the individual performance of the students of $D_2$. Applying equation (2) it is straightforward to check that the mean values of the above TFNs are:

$D_1 = \frac{1}{170}.(60A+40B+20C+30D+20F) \approx (63.53, 73.5, 83.47)$, and $D_2 = \frac{1}{255}.(60A+90B+45C+45D+15F) \approx (65.88, 72.71, 79.53)$.

Therefore, ***Eq. (3)*** gives that $X(D_1) = 73.5$ and $X(D_2) = 72.71$. Consequently, both departments demonstrated a good (C) mean performance, with the performance of $D_1$ being slightly better.

**Example 2.** The individual performance of the five players of a basket-ball team who started a game was assessed by six different athletic journalists using a scale from 0 to 100 as follows: $P_1$ (player 1)**:** 43, 48, 49, 49, 50, 52, $P_2$**:** 81, 83. 85, 88, 91, 95, $P_3$**:** 76, 82, 89, 95, 95, 98, $P_4$**:** 86, 86, 87, 87, 87, 88 and $P_5$**:** 35, 40, 44, 52, 59, 62. It is asked to assess the mean performance of the five players and their overall quality performance by using the linguistic grades A, B, C, D and F. Also, for reasons of comparison, it is asked to approximate their mean performance in two ways, by using TFNs and TpFNs.

- Mean performance: Adding the $5 * 6 = 30$ in total scores assigned by the journalists to the five players and dividing the corresponding sum by 30 one finds that the mean value of those scores is approximately equal to 72.07. Therefore the mean performance of the five players can be characterized as good (C).
- Quality performance: A simple observation of the given data shows that 14 of the 30 in total scores correspond to the linguistic grade A, four to B, one to C, four to D and seven to F. Replacing those values to formula (1) one finds that the GPA index is approximately equal to 2.47. Therefore, the five players' overall quality performance can be characterized as more than satisfactory.

  Using TFNS: Forming the TFNs A, B, C, D and F and observing the $5*6 = 30$ in total player scores it becomes clear that 14 of them correspond to the TFN A, four to B, one to C, four to D and seven to F. The mean value of the above TFNs (***Definition 5*** ) is equal to $M = \frac{1}{30}(14A + 4B + C + 4D + 7F) \approx (58.33, 68.98, 79.63)$.

Therefore the mean performance of the five players is approximated by $X(M) = 68.98$ (good).

- Using TpFNs: A TpFN (denoted, for simplicity, by the same letter) is assigned to each basket-ball player as follows: $P_1 = (0, 43, 52, 59)$, $P_2 = (75, 81, 95, 100)$, $P_3 = (75, 76, 98, 100)$, $P_4 = (85, 86, 88, 100)$ and $P_5 = (0, 35, 62, 74)$. Each of the above TpFNs describes numerically the individual performance of the corresponding player in the form (*a, b, c, d*), where *a* and *d* are the lower and higher scores respectively corresponding to his performance, while *c* and *b* are the lower and higher scores respectively assigned to the corresponding player by the athletic journalists. For example, the performance of the player $P_1$ was characterized by the journalists from unsatisfactory (scores 43, 48, 49, 49) to fair (scores 50, 52), which means that $a = 0$ (lower score for F) and $d = 59$ (lower score for D), etc.

The mean value of the TpFNs $P_i$, i =1, 2, 3, 4, 5 (***Definition 5***), is equal to $P = \frac{1}{5}\sum_{i=1}^{5} P_i = (47, 64.2, 79, 86.6)$.

Therefore, under the light of ***Remark 4*** and ***Proposition 3*** one finds that $X(P) = \frac{2(47+86.6)+7(64.2+79)}{18} \approx 70.53$. This outcome shows that the five players demonstrated a good (C) mean performance.

The outcomes obtained from the application of the assessment methods used in ***Examples 1 & 2*** are depicted in ***Tables (2)-(3)*** below.

***Table 2.*** The outcomes of Example 1.

| **Method** | **$D_1$** | **$D_2$** | **Performance** |
|---|---|---|---|
| **GPA index** | 2.53 | 2.53 | More than satisfactory |
| **TFNs** | 73.5 | 72.68 | Good (C) |

***Table 3.*** The outcomes of Example 2.

| **Method** | | **Performance** |
|---|---|---|
| **Mean value** | 72.07 | Good (C) |
| **GPA index** | 2.47 | More than satisfactory |
| **TFNs** | 68.98 | Good (C) |
| **TpFNs** | 70.53 | Good (C) |

Observing those Tables one can see that the fuzzy outcomes (TFNs/TpFNs) are more or less compatible to the crisp ones (mean value/GPA index). This provides a strong indication that the fuzzy assessment method developed in this work "behaves" well. The appearing, relatively small, numerical differences are due to the different "philosophy" of the methods used (mean and quality performance, bi-valued and fuzzy logic).

The approximation of the player mean performance (70.53) obtained in ***Example 2*** using TpFNs is better (nearer to the accurate mean value 72.07 of the numerical scores) than that obtained by using TFNs (68.98). This is explained by the fact that the TpFNs, due to the way of their construction, describe more accurately than the TFNs each player's individual performance. However, it is not always easy in practice to use TpFNs instead of TFNs. Another advantage of using TpFNs as assessment tools is that, in contrast to TFNs, they make possible the comparison of the individual performance of any two members of the group, even among those whose performance has been characterized by the same qualitative grade. At any case, the fuzzy approximation of a group's mean performance is useful only when no numerical scores are given assessing the idividual performance of its members.

## 4.2. Examples of FLP Problems

**Example 3.** In a furniture factory it has been estimated that the construction of a set of tables needs 2 - 3 working hours (w. h.) for assembling, 2.5 - 3.5 w. h. for elaboration (plane, etc.)

and 0.75 - 1.25 w. h. for polishing, while the construction of a set of desks needs 0.8 - 1.2, 2 - 4 and 1.5 - 2.5 w. h. respectively for each of the above procedures. According to the factory's existing number of workers, at most 20 w. h. per day can be spent for the assembling, at most 30 w. h. for the elaboration and at most 18 w. h. for the polishing of the tables and desks. If the profit from the sale of a set of tables is between 2.7 and 3.3 hundred euros and of a set of desks is between 3.8 and 4.2 hundred euros[2], find how many sets of tables and desks should be constructed daily to maximize the factory's total profit. Express the problem's optimal solution with TFNs of DoF equal to 1.

**Solution.** Let $x_1$ and $x_2$ be the sets of tables and desks to be constructed daily. Then, using TFNs, the problem can be mathematically formulated as follows[3]:

*Maximize* $F = (2.7, 3, 3.3)x_1 + (3.8, 4, 4.2)x_2$ *subject to* $x_1, x_2 \geq 0$ and

$(2, 2.5, 3)x_1 + (0.8, 1, 1.2]x_2 \leq (19, 20, 21),$

$(2.5, 3, 3.5)x_1 + (2, 3, 4)x_2 \leq (29, 30, 31),$

$(0.75, 1, 1.25)x_1 + (1.5, 2, 2.5)x_2 \leq (15, 16, 17).$

The ranking of the TFNs involved leads to the following LP maximization problem of canonical form:

*Maximize* $f(x_1, x_2) = 3x_1 + 4x_2$ *subject to* $x_1, x_2 \geq 0$ and $2.5x_1 + x_2 \leq 20$, $3x_1 + 3x_2 \leq 30$ , and $x_1 + 2x_2 \leq 16$.

Adding the slack variables $s_1$, $s_2$, $s_3$ for converting the last three inequalities to equations one forms the problem's first SIMPLEX matrix, which corresponds to the feasible solution $f(0, 0) = 0$, as follows:

$$\begin{bmatrix} x_1 & x_2 & s_1 & s_2 & s_3 & | & \text{Const.} \\ - & - & - & - & - & - & - \\ 2.5 & 1 & 1 & 0 & 0 & | & 20 = s_1 \\ 3 & 3 & 0 & 1 & 0 & | & 30 = s_2 \\ 1 & 2 & 0 & 0 & 1 & | & 16 = s_3 \\ - & - & - & - & - & | & - \\ -3 & -4 & 0 & 0 & 0 & | & 0 = f(0,0) \end{bmatrix}.$$

Denote by $L_1$, $L_2$, $L_3$, $L_4$ the rows of the above matrix, the fourth one being the *net evaluation row*. Since -4 is the smaller (negative) number of the net evaluation row and $\frac{16}{2} < \frac{30}{3} < \frac{20}{1}$, the *pivot element* 2 lies in the intersection of the third row and second column Therefore, applying the linear transformations $L_3 \rightarrow \frac{1}{2}L_3 = L'_3$ and $L_1 \rightarrow L_1 - L'_3$, $L_2 \rightarrow L_2 - 3L'_3$, $L_4 \rightarrow L_4 + 4L'_3$, one obtains the second SIMPLEX matrix, which corresponds to the feasible solution $f(0, 8) = 32$ and is the following:

$$\begin{bmatrix} x_1 & x_2 & s_1 & s_2 & s_3 & | & \text{Const.} \\ - & - & - & - & - & - & - \\ 2 & 0 & 1 & 0 & -\frac{1}{2} & | & 12 = s_1 \\ \frac{3}{2} & 0 & 0 & 1 & -\frac{3}{2} & | & 6 = s_2 \\ \frac{1}{2} & 1 & 0 & 0 & \frac{1}{2} & | & 8 = x_2 \\ - & - & - & - & - & | & - \\ -1 & 0 & 0 & 0 & 0 & | & 32 = f(0,8) \end{bmatrix}.$$

---

[2]The profit is changing depending upon the price of the wood, the salaries of the workers, etc.

[3] The mathematical formulation of the problem using TFNs is not unique. Here we have taken $b = \frac{a+c}{2}$ for all the TFNs involved, but this is not compulsory. The change of the values of the above TFNs, changes of course the ordinary LP problem obtained by ranking them, but the change of its optimal solution is relatively small.

In this matrix the pivot element $\frac{3}{2}$ lies in the intersection of the second row and first column, therefore working as above one obtains the third SIMPLEX matrix, which is:

$$\begin{bmatrix} x_1 & x_2 & s_1 & s_2 & s_3 & | & \text{Const.} \\ - & - & - & - & - & - & - \\ 0 & 0 & 1 & -\frac{4}{3} & -\frac{3}{2} & | & 4 = s_1 \\ 1 & 0 & 0 & \frac{2}{3} & -1 & | & 4 = x_1 \\ 0 & 1 & 0 & -\frac{1}{3} & 1 & | & 6 = x_2 \\ - & - & - & - & - & | & - \\ 0 & 0 & 0 & \frac{2}{3} & 1 & | & 36 = f(4,6) \end{bmatrix}.$$

Since there is no negative index in the net evaluation row, this is the last SIMPLEX matrix. Therefore $f(4, 6) = 36$ is the optimal solution maximizing the objective function. Further, since both the decision variables $x_1$ and $x_2$ are basic variables, i.e. they participate in the optimal solution, the above solution is unique.

Converting, with the help of ***Proposition 4***, the values of the decision variables in the above solution to TFNs with DoF equal to 1, one finds that $x_1 = (\alpha, 11\text{-}2\alpha, \alpha+1]$ with $\frac{10}{3} < a < \frac{11}{3}$ and $x_2 = (a, 17\text{-}2a, a+1)$ with. $\frac{16}{3} < a < \frac{17}{3}$. Therefore a fuzzy expression of the optimal solution states that the factory's maximal profit corresponds to a daily production between $\alpha$ and $\alpha+1$ groups of tables with $3.33 < a < 3.67$ and between a and $a+1$ groups of desks with $5.33 < a < 5.67$.

However, taking for example $\alpha = 3.5$ and $a = 5.5$ and considering the extreme in this case values of the daily construction of 4.5 groups of tables and 6.5 groups of desks, one finds that they are needed 33 in total w. h. for elaboration, whereas the maximum available w. h. are only 30. In other words, a fuzzy expression of the solution does not guarantee that all the values of the decision variables within the boundaries of the corresponding TFNs are feasible solutions.

**Example 4.** Three kinds of food, say $F_1$, $F_2$ and $F_3$, are used in a poultry farm for feeding the chickens, their cost varying between 38 - 42, 17 - 23 and 55 - 65 cents per kilo respectively. The food $F_1$ contains between 1.5 - 2.5 units of iron and 4 - 6 units of vitamins per kilo, $F_2$ contains 3.2 - 4.8, 0.6 – 1.4 and $F_3$ contains 1.7 – 2.3, 0.8 – 1.2 units per kilo respectively. It has been decided that the chickens must receive at least 24 units of iron and 8 units of vitamins per day. How must one mix the three foods so that to minimize the cost of the food? Express the problem's solution with TpFNs of DoF equal to 2.

**Solution.** Let $x_1$, $x_2$ and $x_3$ be the quantities of kilos to be mixed for each of the foods $F_1$, $F_2$ and $F_3$ respectively. Then, using TpFNs the problem's mathematical model could be formulated as follows[4]:

*Minimize* $F = (38, 39, 41, 42)\, x_1 + (17, 18, 22, 23)\, x_2 + (55, 56, 64, 65]\, x_3$ subject to $x_1, x_2, x_3 \geq 0$ and

$(1.5, 1.8, 2.2, 2.5)\, x_1 + (3.2, 3.5, 4.5, 4.8)\, x_2 + (1.7, 1.9, 2.1, 2.3]\, x_3 \geq [22, 23, 25, 26]$

$[4, 4.5, 5.5, 6]\, x_1 + [0.6, 0.8, 1.2, 1.4]\, x_2 + [0.8, 0.9, 1.1, 1.2] x_3 \geq (6, 7, 9, 1\ 0).$

The ranking of the TpFNs by ***Definition 4*** leads to the following LP minimization problem of canonical form:

*Minimize* $f(x_1, x_2, x_2) = 40x_1 + 20x_2 + 60x_3$ subject to $x_1, x_2, x_3 \geq 0$ and $2x_1 + 4x_2 + 2x_3 \geq 24$, $5x_1 + x_2 + x_3 \geq 8$.

The dual of the above problem is: the following:

*Maximize* $g(z_1, z_2) = 24z_1 + 8z_2$ subject to $t\ z_1, z_2 \geq 0$, $2z_1 + 5z_2 \leq 40$, $4z_1 + z_2 \leq 20$, $2z_1 + z_2 \leq 60$

Working similarly with ***Example 3*** it is straightforward to check that the last SIMPLEX matrix of the dual problem is the following:

---

[4]The problem's mathematical formulation using TpFNs is not unique, but the change of its optimal solution in each case is relatively small.

$$\begin{bmatrix} z_1 & z_2 & s_1 & s_2 & s_3 & | & \text{Const.} \\ - & - & - & - & - & - & - \\ 0 & 1 & \frac{2}{9} & \frac{1}{9} & 0 & | & \frac{20}{3} = z_2 \\ 1 & 0 & -\frac{1}{18} & \frac{5}{18} & 0 & | & \frac{10}{3} = z_1 \\ 0 & 0 & -\frac{1}{9} & -\frac{4}{9} & 1 & | & \frac{140}{3} = s_3 \\ - & - & - & - & - & | & - \\ 0 & 0 & \frac{4}{9} & \frac{52}{9} & 0 & | & \frac{400}{3} = g(\frac{10}{3}, \frac{20}{3}) \end{bmatrix}.$$

Therefore the solution of the original minimization problem is $f_{min} = f(\frac{4}{9}, \frac{52}{9}, 0) = \frac{400}{3}$.

In other words, the minimal cost of the chicken food is $\frac{400}{3} \approx 133$ cents and will be succeeded by mixing $\frac{4}{9} \approx 0.44$ kilos from food $F_1$ and $\frac{52}{9} \approx 5.77$ kilos from food $F_2$.

Converting the values of the decision variables in the above solution to TpFNs with DoF equal to 2 one finds with the help of ***Proposition 5*** that $x_1, x_2, x_3$ must be of the form $(\alpha, b, c, \alpha + 2)$ with

$\alpha < b \leq c < \alpha + 2$, $b + c = \frac{18R - 4a - 4}{7}$ and $R = \frac{4}{9}$ or $R = \frac{52}{9}$ or $R = 0$ respectively.

For R = $\frac{4}{9}$ one finds that b + c = $\frac{4-4a}{7}$. Therefore b < $\frac{4-4a}{7}$ - b or b < $\frac{2-2a}{7}$ which gives that α < $\frac{2-2a}{7}$ or α < $\frac{2}{9}$. Taking for example α = $\frac{1}{9}$, one finds that b < $\frac{2-\frac{2}{9}}{7} = \frac{16}{63}$. Therefore, taking for example b = $\frac{15}{63}$, we obtain that c = $\frac{4-\frac{4}{9}}{7}$ - $\frac{15}{63}$ = $\frac{17}{63}$. Therefore $x_1 = (\frac{7}{63}, \frac{15}{63}, \frac{17}{63}, \frac{133}{63})$.

Working similarly for $R = \frac{52}{9}$ and $R = 0$ one obtains that $x_2 = (\frac{196}{63}, \frac{340}{63}, \frac{362}{63}, \frac{488}{63})$ and $x_3 = (-\frac{21}{63}, -\frac{15}{63}, -\frac{9}{63}, \frac{60}{63})$, respectively.

Therefore, since a TpFN (a, b, c, d) expresses mathematically the fuzzy statement that the interval [b, c] lies within the interval [a, d], a fuzzy expression of the problem's optimal solution states that the minimal cost of the chickens' food will be succeeded by mixing between $\frac{15}{63} \approx 0.24$ , $\frac{17}{63} \approx 0.27$, between $\frac{340}{63} \approx 5.4$ , $\frac{362}{63} \approx 5.75$ and between $-\frac{15}{63} \approx -0.24$ , $-\frac{17}{63} \approx -0.27$ kilos from each one of the foods $F_1$, $F_2$ and $F_3$ respectively. The values of $x_3$ are not feasible and must be replaced by 0, whereas the values of $x_1$ and $x_2$ must be checked as we did in ***Example 3***.

## 5. Discussion and Conclusions

The target of the present paper was two-folded. First, a combination of TFNs / TpFNs and of the COG defuzzification technique was used for assessment purposes. Examples were presented on student and basket-ball player assessment and the new fuzzy method was validated by comparing its outcomes with those of traditional assessment methods (calculation of the mean value of scores and of the GPA index). The advantage of this method is that it can be used for evaluating a group's mean performance when qualitative grades are used instead of numerical scores for assessing the individual performance of its members.

Second, a new technique was developed for solving Fuzzy LP problems by ranking the FNs involved and by solving the ordinary LP problem obtained in this way with the standard theory. Real-life examples were presented to illustrate the applicability of the new technique in practice. In LP problems with a vague structure a fuzzy expression of their solution is often preferable than a crisp one. This was attempted in the present work by converting the values of the decision variables in the optimal solution of the obtained ordinary LP problem to FNs with the desired DoF. The smaller the value of the chosen DoF, the more creditable is the fuzzy expression of the problem's optimal solution.

The new assessment method that has been developed in this work has a general character. This means that, apart for student and athlete assessment, it could be utilized for assessing a great variety of other human or machine (e.g. Case – Based Reasoning or Decision – Making systems) activities. This is an important direction for future research. Also a technique similar to that applied here for solving FLP problems could be used for solving systems of equations with fuzzy coefficients, as well as for solving LP problems and systems of equations with grey coefficients [20].